\documentclass{bmvc2k}

\title{FisheyeDistill: Self-Supervised Monocular Depth Estimation with Ordinal Distillation for Fisheye Cameras}

\addauthor{Qingan Yan}{qingan.yan@innopeaktech.com}{1}
\addauthor{Pan Ji$^*$}{pan.ji@innopeaktech.com}{13}
\addauthor{Nitin Bansal}{nitin.bansal@innopeaktech.com}{1}
\addauthor{Yuxin Ma$^\dagger$}{yuxinma2005@gmail.com}{2}
\addauthor{Yuan Tian}{yuan.tian@innopeaktech.com}{1}
\addauthor{Yi Xu}{yi.xu@innopeaktech.com}{1}

\addinstitution{
 OPPO US Research Center\\
 InnoPeak Technology, Inc.\\
 Palo Alto, California,US
}
\addinstitution{
 Wing LLC\\
 Mountain View, California, US \\
 \small{($^\dagger$Work done while at OPPO US Research Center.)}
}
\addinstitution{
 $^*$Corresponding author
}

\runninghead{Yan \etal}{FisheyeDistill}

\def\eg{\emph{e.g.}} 

\def\ie{\emph{i.e.}}

\def\etal{\emph{et al.}}

\usepackage{bbding}

\begin{document}

\maketitle

\begin{abstract}
In this paper, we deal with the problem of monocular depth estimation for fisheye cameras 
in a self-supervised manner. A known issue of self-supervised depth estimation is that it suffers in low-light/over-exposure conditions and in large homogeneous regions. 
To tackle this issue, we propose a novel ordinal distillation loss that distills the ordinal information from a large teacher model. Such a teacher model, since having been trained on a large amount of diverse data, can capture the depth ordering information well, but lacks in preserving accurate scene geometry. Combined with self-supervised losses, we show that our model can not only generate reasonable depth maps in challenging environments but also better recover the scene geometry. We further leverage the fisheye cameras of an AR-Glasses device to collect an indoor dataset to facilitate evaluation. 
\end{abstract}



\section{Introduction}

Fisheye cameras have gradually gained popularity on Head Mounted Display (HMD), as it offers a  large field-of-view (FoV) that is well suited for immersive AR/VR experience. To estimate the depth map from a distorted fisheye image, a typical prerequisite is to perform image rectification. Such a rectification step, however, undermines the benefits of using fisheye cameras because it will inevitably reduce the FoV of the camera. Moreover, image rectification itself is expensive on HMD as there are only very limited computational resources available on the device. Therefore, it's desirable to directly estimate a depth map from the fisheye image, which is the main theme of this paper.

The past few years have seen a growing interests in self-supervised depth estimation~\cite{garg2016unsupervised, zhou2017unsupervised, godard2019digging} because it requires no groundtruth depth during training and yet achieves comparable results with some of the well known supervised methods. However, due to dearth of  effective benchmarking datasets, depth estimation for fisheye cameras remains under-explored, with only a few exceptions~\cite{kumar2020unrectdepthnet,kumar2020fisheyedistancenet}. Nonetheless, those fisheye depth estimation methods~\cite{kumar2020unrectdepthnet,kumar2020fisheyedistancenet} are only evaluated on well-conditioned outdoor self-driving datasets, including the WoodScape dataset~\cite{yogamani2019woodscape}, for which the groundtruth depths are not yet publicly available. In this work, we follow this trend to take a self-supervised approach to estimating a depth map from a distorted image, but under a more challenging environment, which is captured by HMD.

Self-supervised depth estimation is not without challenges because the key supervision signal of photo-consistency can become ineffective in low-light/over-exposure conditions and in large areas of no/little textures as commonly seen in indoor environments. On the other hand, there has been depth models that are trained on a variety of datasets and can generalize well to new unseen domains~\cite{ranftl2019towards,ranftl2021vision,yin2021learning}. Although such depth models can generate visually good-looking depth maps, the 3D scene geometry in those depth maps are often not well recovered~\cite{yin2021learning}. For example, when visualizing the depth maps of an indoor room in 3D, the horizontal and vertical walls may not be perpendicular to each other, showing distorted 3D scene geometry~\cite{yin2021learning}. 

In this work, we propose to combine the best of self-supervised depth estimation and knowledge distillation~\cite{hinton2015distilling} to build a robust depth model for fisheye cameras especially in challenging indoor environments. In particular, we adapt the photometric loss~\cite{godard2019digging}, which is usually based on a pin-hole camera model, to a fisheye camera model. On top of the self-supervised losses, we further devise a novel ordinal distillation loss which aims to transfer the depth ordering information from a teacher model into our target model. The intuition of ordinal distillation is that the teacher model~\cite{ranftl2019towards,ranftl2021vision} is usually good at predicting the depth ordering relationships of pixels (\eg, a certain pixel is closer or farther than other pixels), albeit not reflecting the exact 3D geometry. Our final depth model becomes more robust and accurate by {\bf (i)} learning the ordering relationships of all pixels via ordinal distillation and {\bf (ii)} respecting the 3D geometry enforced by the photometric loss. 

To perform quantitative evaluation, we collect a new dataset which contains stereo image sequences captured by a pair of stereo fisheye cameras mounted on an AR-Glasses device, and generate the pseudo groundtruth depth maps using stereo matching~\cite{hirschmuller2007stereo}. On this dataset, we show that our proposed model leads to significant improvements over the baseline models. 

\section{Related Work}

In this section, we briefly review a few related works on self-supervised and supervised monocular depth estimation.

\subsection*{Self-Supervised Depth Estimation}
Most of self-supervised depth estimation methods assume that the camera is calibrated and the images are rectified such that a photometric loss can be constructed via a combination of unprojection and projection operations. Garg~\etal~\cite{garg2016unsupervised} are the first to leverage the photo-consistency between stereo images to build a self-supervised loss for training a deep depth model. Zhou~\etal~\cite{zhou2017unsupervised} extend to using temporal sequences to train a depth model and a pose model with a temporal photometric loss. A lot of follow-up methods are then proposed to improve~\cite{garg2016unsupervised,zhou2017unsupervised} by new losses. Depth consistency losses are introduced to enforce the network to predict consistent depth maps in stereo~\cite{godard2017unsupervised} or temporal~\cite{bian2019unsupervised,luo2020consistent} images. Wang~\etal~\cite{wang2018learning} observe a scale diminishing issue in monocular training and propose a depth normalization method to counter this issue. A few methods~\cite{wang2019recurrent,zou2020learning} employ recurrent neural networks to model long-term temporal dependencies for self-supervised training. Some other methods~\cite{chen2019self,yin2018geonet,zou2018dfnet} introduce an additional optical flow network to build cross-task consistencies in a monocular training setup. Godard~\etal~\cite{godard2019digging} comprehensively analyze the challenges faced in self-supervised depth learning, including occlusion, static pixels, and texture-copying artifacts, and propose a set of novel techniques to handle those challenges. Tiwari~\etal~\cite{tiwari2020pseudo} propose to combine self-supervised depth and geometric SLAM in a self-improving loop. Ji~\etal~\cite{ji2021monoindoor} observe the challenges for indoor self-supervised depth estimation and come up with a depth factorization module and a residual pose estimation module to improve the performance under indoor environments. Watson~\etal~\cite{watson2021temporal} use multiple temporal images as input to the depth model and put forth a teacher-student framework to deal with moving objects in the scene. Ji~\etal~\cite{ji2022georefine} propose an online system for refining a depth model in a self-supervised manner, combining a robustified SLAM system and a monocular depth model.


\begin{figure*}[!t]
\begin{center}
\includegraphics[width=1.0\textwidth]{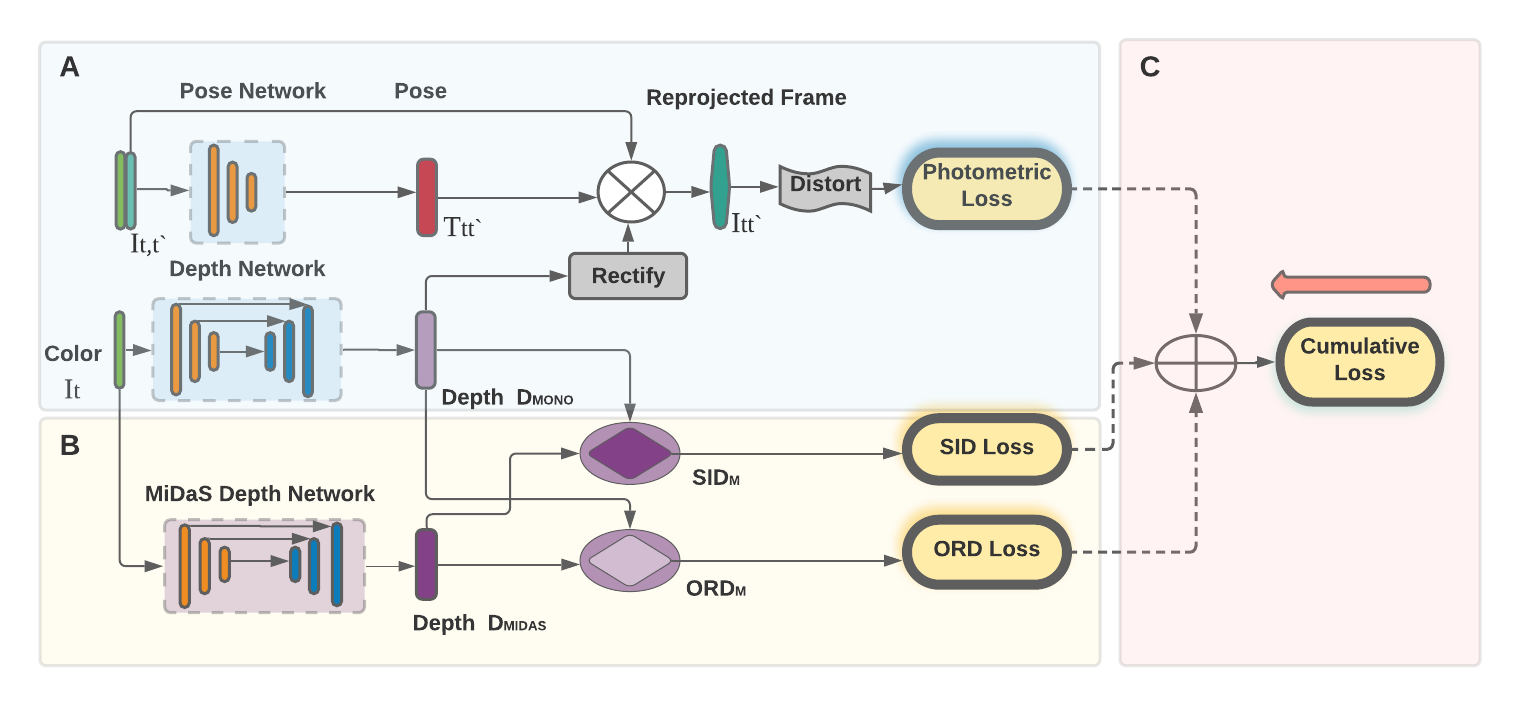}
\end{center}
\vspace{-0.3cm}
\caption{The system workflow of our {\bf FisheyeDistill}. Our system consists of two main parts, \ie, a student model trained with self-supervised losses and distillation losses, and a teacher model (MiDaS)~\cite{ranftl2019towards,ranftl2021vision} that predicts depths to guide the training of the student model. The SID stands for the scale-invariant distillation and $SID_{M}$ is the corresponding function module described in Sec.~\ref{sec:loss}. ORD and $ORD_{M}$ are the ordinal distillation loss and the module.}
\label{fig:workflow}
\end{figure*}

\subsection*{Supervised Depth Estimation}

Ever since EigenDepth~\cite{eigen2014depth}, many methods have been proposed to improve supervised monocular depth estimation on a specific dataset, either by using better loss terms~\cite{fu2018deep,liu2015learning} or via multi-task learning~\cite{ummenhofer2017demon,teed2018deepv2d,zhou2018deeptam,qi2018geonet,guo2018learning,qiao2021vip}. Instead of focusing on the depth prediction of one single dataset (or domain), some other methods seek to train a depth model that can learn across different domains. Along this line, Megadepth~\cite{li2018megadepth} exploits internet photos to train a depth model with structure-from-motion reconstructions~\cite{schonberger2016structure}. Li~\etal~\cite{li2019learning} leverage internet videos with frozen people to predict sharp depth maps for images with people. MiDaS~\cite{ranftl2019towards} proposes a scale-invariant loss to train the depth model on a large and diverse training sets, which facilitates generality and cross-dataset learning  by providing a dataset-agnostic depth model.                            DPT~\cite{ranftl2021vision} further improves MiDaS by using a vision transformer network~\cite{dosovitskiy2020image}. Although visually pleasing depth maps can be generated by those models~\cite{ranftl2019towards,ranftl2021vision}, their quantitative performance on a specific dataset is usually worse than those that are trained specifically on the same dataset. This means that the accurate 3D geometry information is not well preserved in those models~\cite{ranftl2019towards,ranftl2021vision}.

Our method tries to reap the benefits of both self-supervised and supervised models in the sense that it respects the 3D geometry via the use of self-supervised losses and enforces ordering consistency between neighboring depth pixels by distilling information from a diversely trained teacher model.


\section{Method}

In this work, we aim at learning a self-supervised monocular depth estimation model for fisheye images (see Fig.~\ref{fig:workflow}). In all the previous works~\cite{garg2016unsupervised,zhou2017unsupervised,godard2019digging}, view synthesis is usually used in self-supervision by learning a depth and pose relationship to synthesize virtual target images from neighboring views $I_{t-1}$ and $I_{t+1}$. For this, it requires a projection function $\Phi$ that maps 3D points $P_i$ in 3D space to image coordinates $p_i = \Phi(P_i)$, and accordingly the corresponding unprojection function $\Phi^{-1}$, which converts image pixels, based on the estimated depth map $D$, into 3D space $P_i = \Phi^{-1}(p_i, D)$ in order to acquire color information from other views.

\subsection{Fisheye Geometry Model}

For fisheye cameras, given a 3D point $P_i = (X_i, Y_i, Z_i)^T$ in camera coordinates and with $(x_i = X_i / Z_i, y_i = Y_i / Z_i)$, the projection function $p_i = \Phi(P_i)$ from 3D point $P_i$ to distorted image pixel $p_i = (u, v)^T$ can be obtained through the following mapping equation:
\begin{equation}
\label{eq:distortion}
\Phi(P_i)=
\begin{bmatrix}
u = f_x \cdot x_d + c_x \\
v =  f_y \cdot y_d + c_y
\end{bmatrix}
,
\end{equation}
where $(x_d = \dfrac{x_i}{r}\phi_{\theta} , y_d = \dfrac{y_i}{r}\phi_{\theta})$ with $r = \sqrt{{x_i}^2 + {y_i}^2}$, $\theta = \arctan(r)$ is the angle of incidence, $\phi_{\theta} = \theta \cdot (1 + k_1 \cdot \theta^2 + k_2 \cdot \theta^4 + k_3 \cdot \theta^6 + k_4 \cdot \theta^8)$ is the polynomial radial distortion model mapping the incident angle to the image radius, $(f_x, f_y)$ and $(c_x, c_y)$ stand for the focal length and the principal point derived from the intrinsic matrix $K$, and $\{k_i\}$ denote the set of fisheye distortion coefficients.

For the unprojection part, due to the existence of fisheye distortion, we are unable to directly transform a pixel $p_i$ into camera coordinates via the pinhole model. In order to achieve that, we first generate an intermediate rectified depth map $\hat{D}$ from the depth map $D$ estimated from our network. This can be done by warping a pixel grid according to Eq.~\eqref{eq:distortion}. Then we leverage the rectified $\hat{D}$ to unproject the grid into 3D by simply applying $\Phi^{-1}(p_i, \hat{D}) = \hat{D} K^{-1} \cdot p_i$.~\footnote{Here and below, we omit a necessary step of converting to the homogeneous coordinates for notation simplicity.} The view synthesis process in our fisheye pipeline can be summarized as: (i) Unproject a uniform pixel grid which has the same resolution with input frames through $\Phi^{-1}(p_i, \hat{D})$; (ii) Project those 3D points by $\Phi(P_i)$ and the associated pose information from pose network module, to obtain distorted synthesis images. This process is different from the FisheyeDistanceNet~\cite{kumar2020fisheyedistancenet} which needs to numerically calculate $\theta$ out from the $4^{th}$ order polynomial.

\subsection{Self-Supervised Losses} 

Following~\cite{godard2019digging}, we use a photometric reprojection loss between two neighboring fisheye images and an edge-aware depth smoothness loss as the self-supervised losses to train the models. Specifically, given a target fisheye image $I_t$ and a source fisheye image $I_t'$, two models (\ie, a depth model and a pose model) are jointly trained to predict a dense depth map $D_t$ and a relative transformation matrix $T_{tt'}$. The per-pixel minimum photometric reprojection loss~\cite{godard2019digging} can be computed as
\begin{equation}
    L^{ph} = \min\limits_{t'} \rho(I_t, I_{tt'}),
\end{equation}
and
\begin{equation}
    I_{tt'} = I_{t'}\big\langle \Phi\big(T_{tt'} \Phi^{-1}(p_i, \hat{D})\big) \big\rangle,
\end{equation}
where $t' \in \{t-1, t+1\}$, $\rho(\cdot)$ denotes a weighted combination of the L1 and Structured SIMilarity (SSIM) losses~\cite{godard2019digging}, and $\big\langle \cdot \big\rangle$ is the bilinear sampling operator. The edge-aware smoothness loss is defined as
\begin{equation}
    L^{sm} = |\partial_{x}d^{\ast}_{t}|e^{-|\partial_{x}I_t|}+|\partial_{y}d^{\ast}_{t}|e^{-|\partial_{y}I_t|},
\end{equation}
where $d^{\ast}_t = d/\bar{d}_t$ is the mean-normalized inverse depth from~\cite{wang2018learning}. An auto-masking mechanism~\cite{godard2019digging} is also adopted to mask out static pixels when computing the photometric loss.

\subsection{Distillation Losses}
\label{sec:loss}

Self-supervised losses are volatile when the brightness or RGB values of a pixel become indistinguishable with its neighboring pixels. Previously, such color ambiguity is implicitly handled via the smoothness loss and a multi-scale strategy~\cite{godard2019digging}, but they do not work well if the textureless regions are large. For example, in indoor environments, textureless walls often occupy a large area in the image; or in low-light/over-exposure conditions, most of the regions in an image are near-textureless. To provide supervision in those scenarios, we apply knowledge distillation~\cite{hinton2015distilling} to distill depth information from a diversely trained teacher model~\cite{ranftl2019towards,ranftl2021vision} into a smaller target model. In particular, we propose to combine a novel depth ordinal distillation loss and a scale-invariant distillation loss~\cite{ranftl2019towards}, which we detail as below.


Pretrained models such as MiDaS~\cite{ranftl2019towards,ranftl2021vision} are good at predicting relative depth relationships as they are trained on large and diverse datasets. In light of this, we aim to distill from the teacher model the ordinal information between neighboring pixels into our student model. Inspired by~\cite{chen2016single}, we propose to use a ranking loss to build an ordinal distillation loss between the depth maps from the teacher model and the target student model.

Given a depth map $D^{MONO}$ from the student model and a depth map $D^{MIDAS}$ from the teacher model, we compute a ranking loss from each pixel $I(i,j)$ to its left neighbor $I(i-1,j)$ as follows,
\begin{equation}
    L_l^{od} = \begin{cases}
    \log\big(1 + \exp(-D_{i,j}^{MONO} + D_{i-1,j}^{MONO})\big), & \text{if $D_{i,j}^{MIDAS} > \alpha D_{i-1,j}^{MIDAS}$} \\
    \log\big(1 + \exp(D_{i,j}^{MONO} - D_{i-1,j}^{MONO})\big), & \text{if $D_{i,j}^{MIDAS} < \beta D_{i-1,j}^{MIDAS}$} \\
    |D_{i,j}^{MONO} - D_{i-1,j}^{MONO}|, & \text{otherwise,}
    \end{cases}
    \label{eq:distill-left}
\end{equation}
where $\alpha$ and $\beta$ are hyper-parameters controlling the ranking gap and are empirically set to $\alpha=1.1$ and $\beta=0.9$ in our experiments. Similarly, we can compute a ranking loss from each pixel $I(i,j)$ to its neighbor above it: $I(i,j-1)$ as $L_t^{od}$ . The final ordinal distillation loss is then the sum of $L_l^{od}$ and $L_t^{od}$, \ie,
\begin{equation}
    L^{od} = L_l^{od} + L_t^{od}.
\end{equation}

Intuitively, if a pixel is predicted to be farther (or closer) than its neighbors by the teacher model, by minimizing the ordinal distillation loss, the student model is encouraged to predict similar depth ordering relationships. If in the teacher model two neighboring pixels are predicted to be close, a smoothness term is enforced for the student model.

We further add a scale-invariant distillation loss $L^{sd}$ between $D^{MONO}$ and $D^{MIDAS}$ to strengthen the supervision in textureless regions. Following~\cite{ranftl2019towards}, $D^{MONO}$ and $D^{MIDAS}$ are respectively normalized to be scale- and shift-invariant (subtracted by the median scale and then divided by the mean shift). The normalized versions are denoted as $\tilde{D}^{MONO}$ and $\tilde{D}^{MIDAS}$. The scale-invariant loss is then defined as:
\begin{equation}
    L^{sd} = |\tilde{D}^{MONO} - \tilde{D}^{MIDAS}|.
\end{equation}

Our final training loss is a weighted combination of the photometric loss, the smoothness loss and the distillation losses, \ie,
\begin{equation}
L = L^{ph} + w^{sm} L^{sm} + w^{od} L^{od} + w^{sd} L^{sd}.
\label{eq:total-loss}
\end{equation}


\section{AR-Glasses Fisheye Dataset}
In order to evaluate the model, we contribute a large fisheye dataset for training and evaluation.
The dataset contains $321,300$ raw fisheye images collected from a well calibrated AR-Glasses device in a stereo manner. All images are acquired in an indoor environment with $8$ different scenes. Besides various chairs, desks and decorations, there are also many lights, texture-less walls and glasses therein. In addition, due to the computational limit on the headset, all frames are captured under the resolution of $640 \times 400$ in gray-scale with a sub-optimal auto-exposure mechanism. These make our 
dataset very challenging for the task of self-supervised depth estimation as the photometric constraints would act poorly in such scenes. We will have more discussion on it in Sec.~\ref{sec:eval}.

More specifically, in our experiments, we use $238,186$ images for training and $42,688$ and $20,213$ images as the validation and test sets respectively. The $8$ scenarios are randomly shuffled and fed into the network while training. Similar to~\cite{godard2019digging}, we also apply the static frame filtering to clean our data and get a final training and a validation set of $71,688$ and $6,265$ images. One major limitation of other fisheye datasets~\cite{matsuki2018omnidirectional} is that there is no easily accessible ground truth for evaluation. Instead, we utilize the semi-global stereo matching method~\cite{hirschmuller2007stereo} based on OpenCV's implementation to calculate pseudo ground truth for our dataset. In our experiments, we select $300$ well calculated stereo depth maps from different viewpoints as ground truth. We conduct all the experiments in the monocular mode.

\section{Experiments}

\subsection{Implementation Details}

We implement our system using Pytorch and employ Adam optimizer to minimize the objective function in Eq.~\eqref{eq:total-loss}. We train the model for 20 epochs with a learning rate of $10^{-4}$ on a 12GB GeForce RTX. Due to the limitation of graphic memory, a small batch size of 6 is leveraged while training and batch normalization~\cite{ioffe2015batch} layers are also frozen to avoid unexpected degeneration. We convert the sigmoid output $\sigma$ from the network to depth via $D = 1 / (a \cdot \sigma + b)$, where $a$ and $b$ are chosen to bound $D$ between $0.1$ and $100$ units. As the resolution of our AR-Glasses images is $640 \times 400$, we feed $640 \times 384$ pixels as the network input to approximate the original aspect ratio. The smoothness loss weight $w^{sm}$ is set to $10^{-3}$ and distillation loss weights $w^{od} = w^{sd} = 1.0$. For generating the rectified intermediate depth $\hat{D}$, we apply a sightly small focal length than the one given in $K$. This would help the network to better learn the content on distorted image boundaries. In all our experiments, we use $0.8$ as the scale factor.

As previously discussed, while the MiDaS teacher model~\cite{ranftl2021vision} provides a good guidance for depth estimation even on challenging structures, it suffers from the limitation in keeping true geometric relationship, due to the lack of cross-view photometric constraints. Therefore, directly adopting a constant weight for the knowledge distillation through all epochs would weaken the ability of learning better geometry information. To circumvent this issue, we adopt a decay weighting strategy for the distillation losses in training. After each $10k$ steps, we decay the weights $w^{od}$ and $w^{sd}$ by applying a factor $s = 0.9^{2 \lambda}$, where $\lambda = \dfrac{steps}{10000}$. Such a strategy enables a major contribution from the distillation losses when the uncertainty is high and facilitates the convergence of photometric constraints in the later training stage. After five epochs, photometric losses begin to play a more dominant role in model training, which ensures the acquisition of geometric relationship between temporal frames. Similar to other self-supervised works, we also set the length of the training snippets to 3 and build the losses over 4 image scales. 

\subsection{Evaluation}
\label{sec:eval}

We evaluate our model mainly on the proposed AR-Glasses fisheye dataset. We do not leverage KITTI~\cite{geiger2013vision} as it is not a fisheye dataset. Woodscape~\cite{yogamani2019woodscape} dataset could be a good candidate, but it is still not fully released. Currently only around $8k$ images are available and without ground truth for evaluation. While the image resolution of AR-Glasses dataset is smaller than KITTI dataset, the ground truth is much denser. Moreover, since there is no explicit sky region in the indoor environment, we do not perform any cropping for each test image. We evaluate the models on all pixels whose depth values are not zero in the computed groundtruth.

The quantitative results are reported in Table~\ref{tab:eval}, which shows the performance of our model on the AR-Glasses dataset. From the table, we can see that without considering the fisheye distortion, Monodepth2~\cite{godard2019digging} is unable to extract plausible depth outputs from distorted images. MiDaS~\cite{ranftl2021vision} shows a good generalization ability on our dataset, despite being trained on rectified images. We are unable to compare with the fisheye model~\cite{kumar2020fisheyedistancenet}, as its source code 
is not publicly available. As fisheye cameras are designed for near-field sensing and deployed on AR-Glasses for indoor applications, the groundtruth depth values are capped at $20m$ for evaluation. 
In addition, we also demonstrate how the decay factor $s$ affects the performance in training. We report the comparison results by employing $s = 0.9^{2 \lambda}$ and $\hat{s} = 0.98^{2 \lambda}$ respectively, where $\hat{s}$ enforces more emphasis on the distillation part. It can be found that letting the network learn more geometric information through the temporal photometric loss helps to improve the results.

\begin{figure}[!t]
\begin{center}
\includegraphics[width=1.0\textwidth]{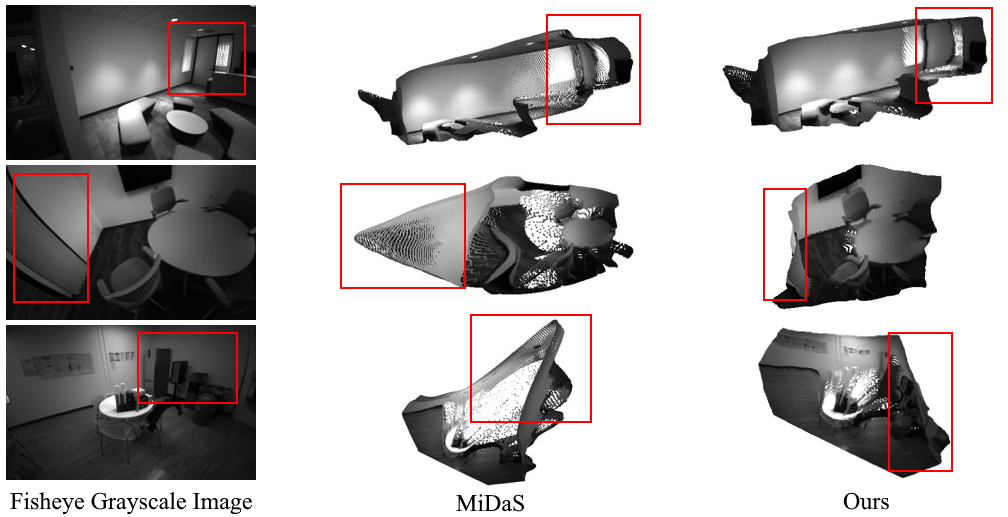}
\end{center}
\caption{Visual comparison of the 3D point clouds generated from depth maps by MiDaS (middle) and our method (right). Compared to MiDaS, our results generate more geometrically correct point clouds.}
\label{fig:vis_pc}
\end{figure}

We also render the qualitative results to demonstrate our advantages in challenging indoor environments, as shown in Fig.~\ref{fig:vis_euroc}. Directly training upon photometric constraints without knowledge distillation is prone to producing halo and texture-copy artifacts. This is because that large texture-less continuous surfaces, such as planar white walls and desks, and the low light condition make photometric losses unable to well establish pixel correspondences from the very beginning. 
In contrast, due to the good generalization ability on different datasets, the MiDaS teacher model~\cite{ranftl2021vision}, provides a smooth and sharp depth ordering hint to let the network approximate plausible depth even on texture-less planes. However, the geometric constraint within MiDaS is insufficient. This is notable from Fig.~\ref{fig:vis_pc}, where we visualize the 3D geometry unprojected from corresponding depth maps of different methods. While the depth map is visually-pleasant, MiDaS suffers from severe distortion in 3D space. Our method overcomes the limitations existing in both~\cite{godard2019digging} and~\cite{ranftl2021vision} and is able to give more consistent results on the challenging indoor dataset.

\begin{table}[t]
\label{tab:eval}
\footnotesize
\centering
\caption{Quantitative results of different algorithms on AR-Glasses fisheye dataset. All the approaches are evaluated in the monocular mode and the estimated depths are scaled using median scaling.}
\vspace{0.2cm}
\begin{tabular}{llllllll}
\hline
Method & Abs Rel & Sq Rel & RMSE & RMSE\_{log} & $\delta < 1.25$ & $\delta < 1.25^2$ & $\delta < 1.25^3$ \\ \hline \hline
Monodepth2~\cite{godard2019digging}  & 0.348 & 0.108 & 0.198 & 0.398 & 0.449 & 0.749 & 0.895 \\
MiDaS~\cite{ranftl2021vision} & 0.216 & 0.131 & 0.252 & 0.267 & 0.756 & 0.918 & 0.965 \\ \hline
Ours with $s = 0.98^{2 \lambda}$  & 0.136 & 0.022 & 0.103 & 0.183 & 0.843 & 0.971 & 0.991 \\
Ours with $s = 0.9^{2 \lambda}$ & 0.131 & 0.020 & 0.098 & 0.177 & 0.854 & 0.973 & 0.992 \\ \hline
\end{tabular}
\end{table}

\begin{figure}[!t]
\begin{center}
\includegraphics[width=0.24\textwidth]{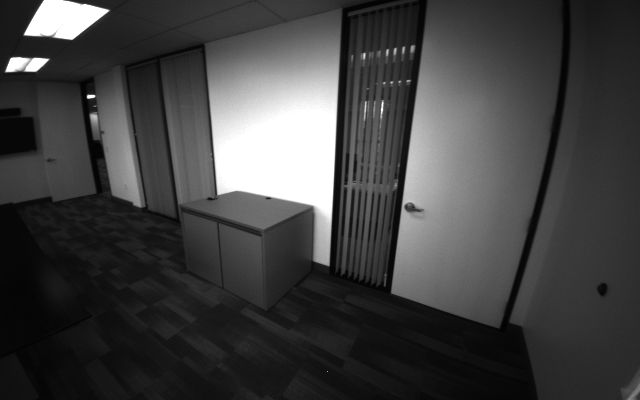}
\includegraphics[width=0.24\textwidth]{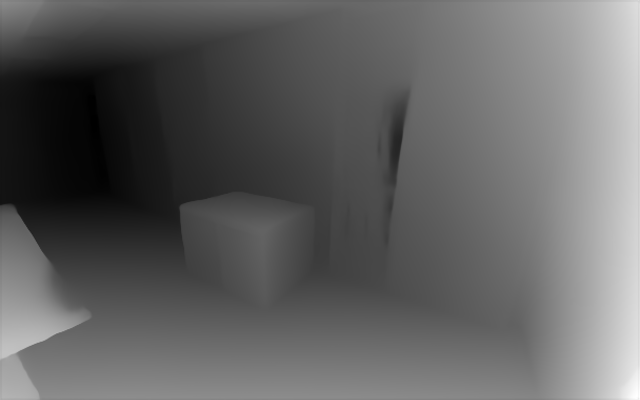}
\includegraphics[width=0.24\textwidth]{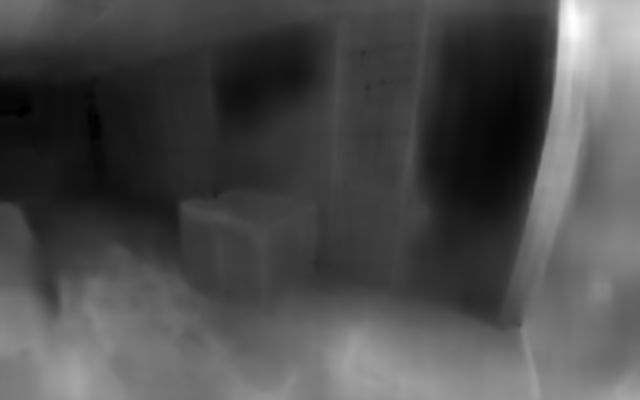} 
\includegraphics[width=0.24\textwidth]{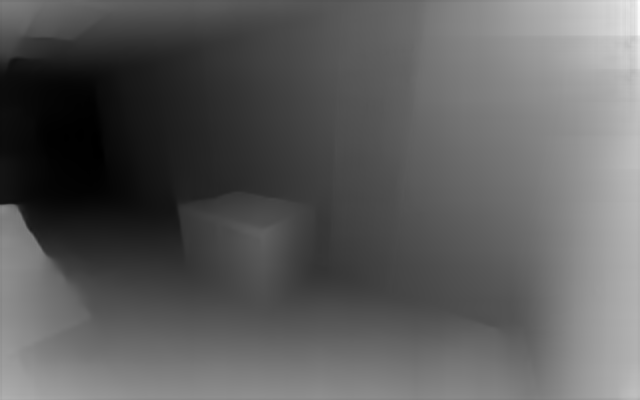} \\
\includegraphics[width=0.24\textwidth]{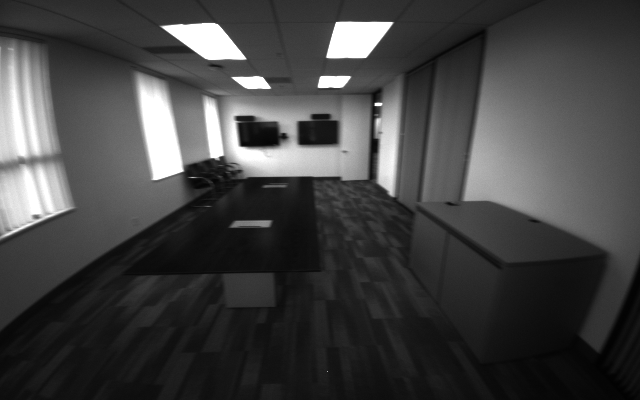}
\includegraphics[width=0.24\textwidth]{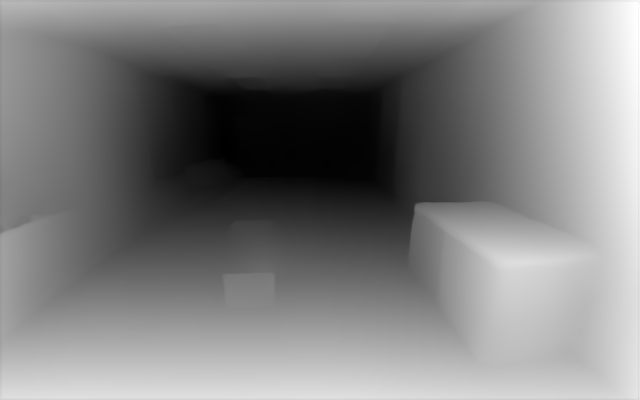}
\includegraphics[width=0.24\textwidth]{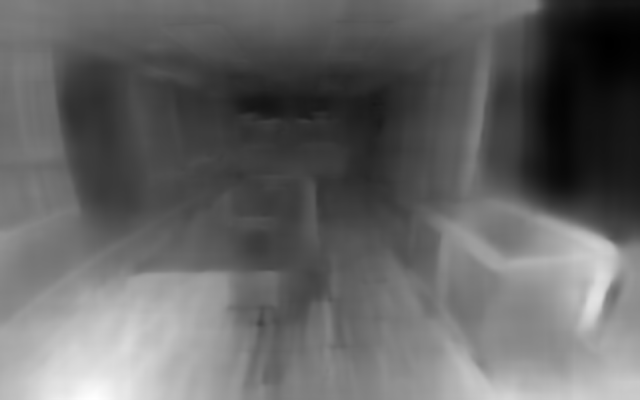}
\includegraphics[width=0.24\textwidth]{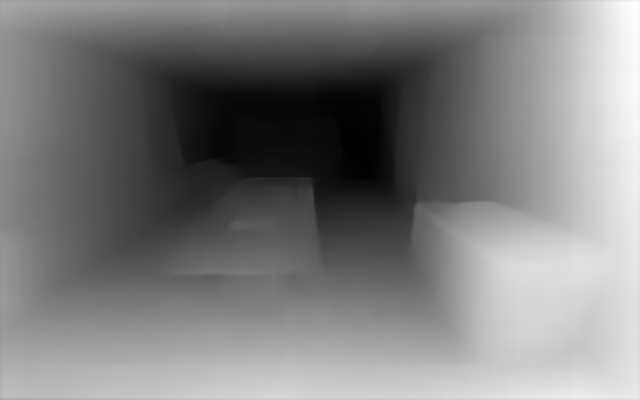} \\
\includegraphics[width=0.24\textwidth]{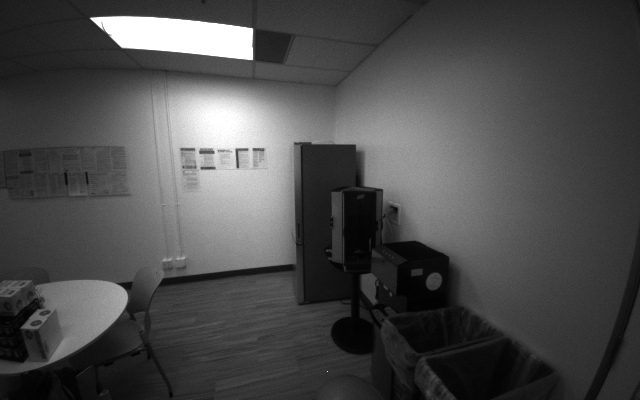}
\includegraphics[width=0.24\textwidth]{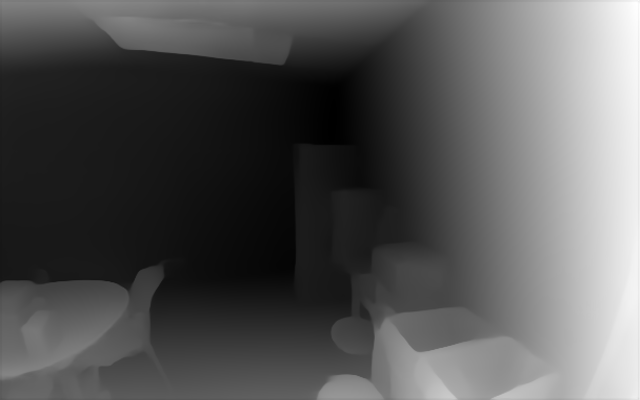}
\includegraphics[width=0.24\textwidth]{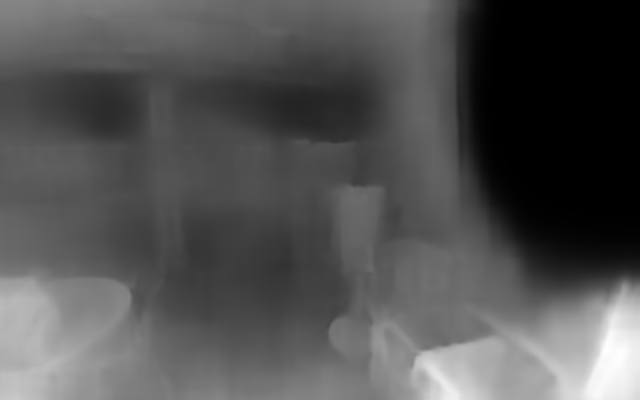}
\includegraphics[width=0.24\textwidth]{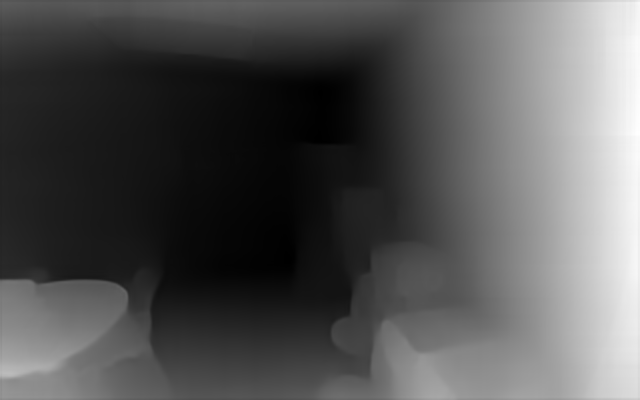}
\end{center}
\caption{Qualitative results on the AR-Glasses dataset. From left to right, the images are respectively input fisheye images, MiDaS estimated disparity maps, disparity maps from our network without any distillation losses, and disparity maps with distillation losses.}
\label{fig:vis_euroc}
\end{figure}

\begin{table}[t]
\label{tab:ablation}
\scriptsize
\centering
\caption{Ablation study by using different variants on the proposed AR-Glasses dataset. Depths are all capped at $20 m$. SID, ORD, DW respectively represent the scale-invariant distillation, ordinal distillation and decay weighting.}
\vspace{0.2cm}
\begin{tabular}{lllllllllll}
\hline
Method & SID & ORD & DW & Abs Rel & Sq Rel & RMSE & RMSE\_{log} & $\delta < 1.25$ & $\delta < 1.25^2$ & $\delta < 1.25^3$ \\ \hline \hline
Ours  & \XSolidBrush & \XSolidBrush & \XSolidBrush & 0.151 & 0.021 & 0.103 & 0.200 & 0.780 & 0.961 & 0.992 \\
Ours & \Checkmark & \Checkmark & \XSolidBrush & 0.134 & 0.020 & 0.098 & 0.180 & 0.846 & 0.972 & 0.992 \\
Ours & \Checkmark & \XSolidBrush & \Checkmark & 0.134 & 0.019 & 0.097 & 0.179 & 0.844 & 0.972 & 0.992 \\
Ours & \Checkmark & \Checkmark & \Checkmark & 0.131 & 0.020 & 0.098 & 0.177 & 0.854 & 0.973 & 0.992 \\ \hline
\end{tabular}
\end{table}

\subsection{Ablation Study}

We conduct an ablation study to illustrate the benefits of different components used in our system and show the results in Table~\ref{tab:ablation}. (i) Remove both distillation losses: The network is trained only with distortion-based self-supervised photometric losses. As previously discussed, the proposed AR-Glasses dataset is very challenging, due to under-exposure and textureless structures. Only relying on photometric constraints is prone to producing halo and texture-copy artifacts. In contrast, the distillation strategy offers a huge quantitative improvement. Fig.~\ref{fig:vis_pc} also provides a visual comparison for the sake of illustration. (ii) Remove the ordinal distillation loss: Ordinal distillation loss is good at the supervision in textureless regions. Removing the ordinal distillation loss from the network would diminish the performance gain over the baseline. (iii) Utilize constant distillation loss weights: Apart from the comparison of using different $s$ as illustrated before, we also launch an experiment with a constant distillation weighting strategy. For the trade-off between keeping depth ordering and learning true geometry, we set the loss weights to $w^{od} = w^{sd} = 0.3$ without decaying here for testing. It can be seen that using constant distillation loss weights decrease the performance on our dataset, which verifies the effectiveness of our decay-weighting strategy.

\section{Conclusions}
In this work, we have presented FisheyeDistill, a self-supervised monocular depth estimation method for fisheye cameras. Our key insight is to combine self-supervised fisheye photometric losses with a novel ordinal distillation loss to train a robust depth model that can work well in challenging environments for fisheye cameras.

\section*{Appendix}

We adopt the same network structures with~\cite{godard2019digging}. In all experiments we use a standard ResNet18~\cite{he2016deep} encoder for both depth and pose networks. The pose encoder takes fisheye frames as input, as our network is trained to learn the pose under the distortion circumstance. We also empirically found that using corresponding rectified images as input to the pose network decreases its performance. For the distillation part, we leverage the pretrained DPT~\cite{ranftl2021vision} large model as our teacher model. For further demonstration, we also upload a demo video, which respectively shows a portion of our contributed AR-glass fisheye dataset and the corresponding depth estimation results (with and without knowledge distillation). It is evident that without the proposed distillation strategies, the results suffer from significant halo artifacts, as is also shown below.

\begin{figure}[h]
\begin{center}
\includegraphics[width=\textwidth]{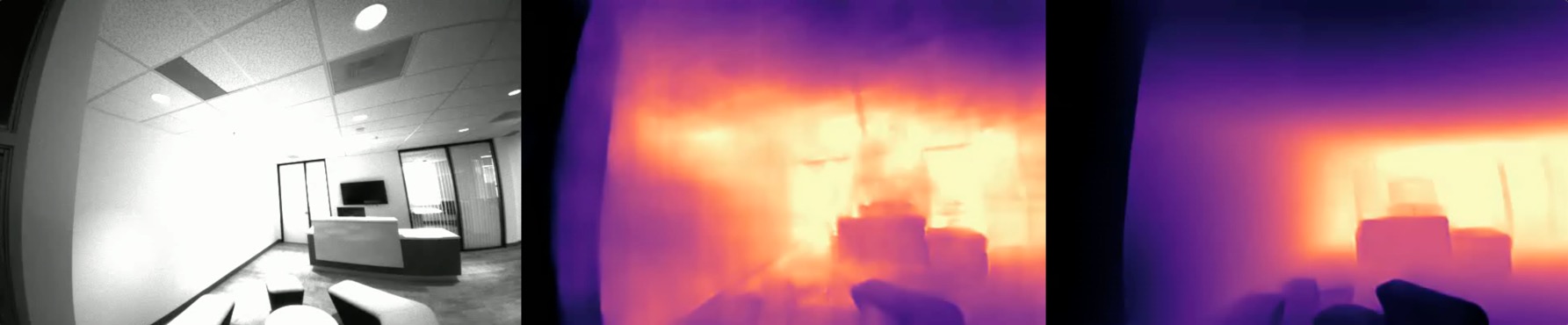}
\includegraphics[width=\textwidth]{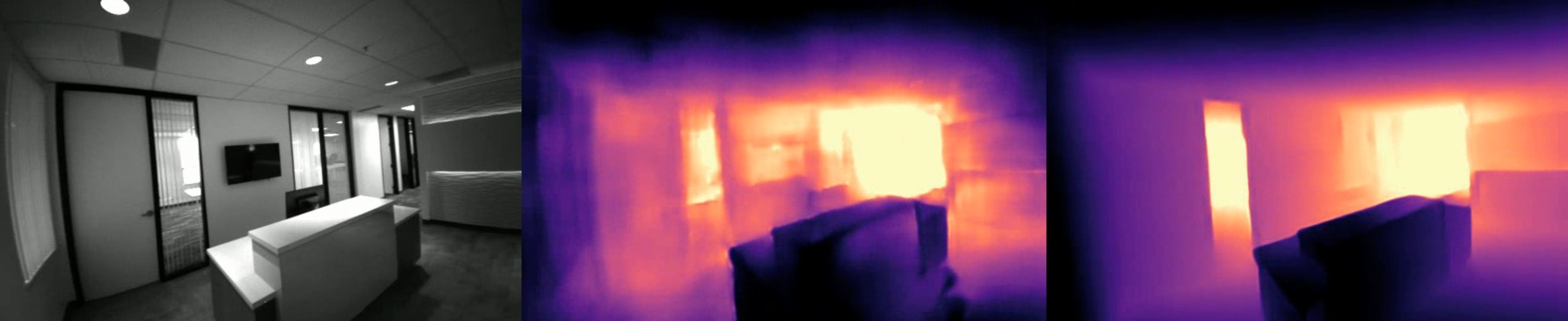}
\end{center}
\caption{Depth estimation results on the AR-Glasses dataset. 
From left to right are respectively the fisheye frames in our dataset, and the color-coded depth predictions from our network without and with using the proposed distillation strategies.
}
\label{fig:vis_euroc}
\end{figure}

\bibliography{egbib}
\end{document}